\pdfoutput=1

\documentclass[11pt]{article}

\usepackage[final]{acl}

\usepackage{times}
\usepackage{latexsym}

\usepackage[T1]{fontenc}

\usepackage{listings}
\usepackage[utf8]{inputenc}

\usepackage{microtype}

\usepackage{inconsolata}

\usepackage{graphicx}

\usepackage{amsmath}
\usepackage{amssymb}
\usepackage{adjustbox}
\usepackage{microtype}
\usepackage{float}
\usepackage{hyperref}
\usepackage{url}
\usepackage{booktabs}
\usepackage{geometry}
\usepackage{multirow}
\usepackage{algorithm}
\usepackage{algorithmic}
\usepackage{wrapfig}
\usepackage{subcaption}
\usepackage{amsmath}
\usepackage{listings}
\usepackage{natbib}

\def\D{{\bf D}}

\def\f{{\bf f}}
\def\F{{\bf F}}

\def\k{{\bf k}}

\def\X{{\bf X}}

\def\x{{\bf x}}

\def\0{{\bf 0}}
\def\1{{\bf 1}}

\def\argmin{\mathop{\rm argmin}}

\title{RAFT: Realistic Attacks to Fool Text Detectors}
\author{James Wang$^1$\thanks{These authors contributed equally to the work}, Ran Li$^1$\footnotemark[1], Junfeng Yang$^1$, Chengzhi Mao$^{2}$\\
Columbia University$^1$, Rutgers University$^2$ \\
\texttt{\{jlw2247, rl3424, jy2324\}@columbia.edu, cm1838@rutgers.edu}
}

\begin{document}
\maketitle

\vspace{-4.5em}

\begin{abstract}
Large language models (LLMs) have exhibited remarkable fluency across various tasks. However, their unethical applications, such as disseminating disinformation, have become a growing concern. Although recent works have proposed a number of LLM detection methods, their robustness and reliability remain unclear. In this paper, we present \emph{RAFT}: a grammar error-free black-box attack against existing LLM detectors. In contrast to previous attacks for language models, our method exploits the transferability of LLM embeddings at the word-level while preserving the original text quality. We leverage an auxiliary embedding to greedily select candidate words to perturb against the target detector. Experiments reveal that our attack effectively compromises all detectors in the study across various domains by up to 99\%,  and are transferable across source models. Manual human evaluation studies show our attacks are realistic and indistinguishable from original human-written text. We also show that examples generated by \emph{RAFT} can be used to train adversarially robust detectors. Our work shows that current LLM detectors are not adversarially robust, underscoring the urgent need for more resilient detection mechanisms.
\end{abstract}
\section{Introduction}
Large language models (LLMs) such as ChatGPT \citep{chatgpt}, LLaMA \cite{touvron2023llama}, and GPT-4 \cite{achiam2023gpt} have exhibited transformative abilities to generate remarkably fluent and cogent long-form text in response to user queries. However, LLMs have been misused to disseminate disinformation, commit academic dishonesty, and launch targeted spear phishing campaigns against vulnerable populations \citep{hazell2023large}. To mitigate harm from malicious use, the capability to distinguish machine-generated text and human-written text is paramount. 

To defend against these malicious use cases, various methods have been developed to successfully detect machine-generated text such as OpenAI's GPT-2 supervised detector \citep{solaiman2019release}, watermarking \citep{kirchenbauer2023watermark}, and likelihood-based zero-shot detectors such as DetectGPT \citep{mitchell2023detectgpt}. In response, red-teaming methods for attacking machine-generated text detectors were created to identify vulnerabilities. Red-teaming methods are primarily based on paraphrasing or word substitution. Paraphrasing-based attacks such as \emph{DIPPER} fine-tune a generative language model on a large set of manually collected paraphrase pairs \citep{krishna2024paraphrasing,sadasivan2023can}. Word substitution-based attacks have leveraged masked language models or auxiliary LLMs to generate replacement candidates \citep{krishna2024paraphrasing, shi2024red}. Despite their effectiveness in subverting various detectors, word substitution-based attacks often contain numerous grammatical errors and semantic inconsistencies that are readily discernible upon human evaluation. 

\begin{figure*}[h]
    \includegraphics[width=\textwidth]{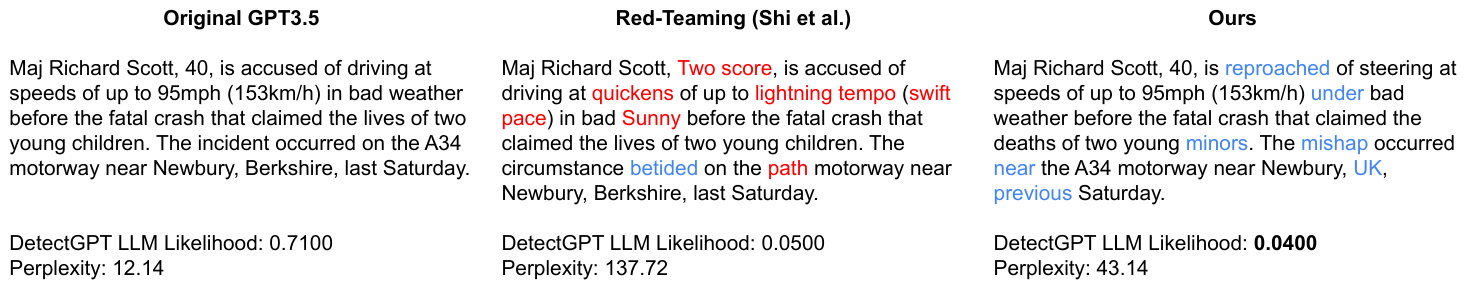}
    \caption{\emph{RAFT} can attack a sample text generated by GPT-3.5-turbo more effectively to subvert detection by DetectGPT than recent red teaming attack efforts \cite{shi2024red} while preserving language fluency and semantic consistency. By enforcing grammatical consistency in the substituted words through POS correction, \emph{RAFT} achieves significantly lower perplexity than attacks that do not enforce grammar. Qualitative evaluation also highlight \emph{RAFT}'s language fluency and semantic consistency with the original text. \textcolor{red}{Red} text represents substituted words with grammatical errors or semantic inconsistencies. \textcolor{blue}{Blue} text represent error-free substitutions.}
    \label{fig:teaser}
    \vspace{-0.85em}
\end{figure*}

In this work, we explore whether machine-generated text can subvert detection with \textit{realistic} perturbations that remain inconspicuous to human readers. A perturbation is considered realistic if it maintains part-of-speech (POS) consistency, minimally increases perplexity, and is indistinguishable from human-written text in manual evaluations.

We present \emph{RAFT}, a zero-shot black-box attack framework to subvert machine-generated text detectors. \emph{RAFT} leverages an auxiliary LLM embedding to optimally select words in machine-generated text for substitution by performing a proxy task. It then employs a black-box LLM to generate replacement candidates, greedily selecting the one that most effectively subverts the target detector. \emph{RAFT} only requires access to an LLM's embedding layer, making it easily deployable and adaptable with the numerous powerful open-source LLMs available \citep{huggingface}.

Our results show \emph{RAFT} can reduce detection performance by up to 99\% while preserving the part-of-speech and semantic consistency of the replaced words. Additionally, we show that the replacement words selected to greedily subvert one target detector can be effectively transferred to attack other detectors, outperforming benchmarked attack methods. Furthermore, we demonstrate that \emph{RAFT}'s outputs can be leveraged to enhance a detector's robustness through adversarial training. Our findings suggest that current detectors are vulnerable to adversarial attacks and highlight the urgency to develop more resilient detection mechanisms. Our code and data is available at \url{https://www.github.com/jameslwang/raft}. 
\begin{figure*}[h]
    \centering
    \includegraphics[width=\textwidth]{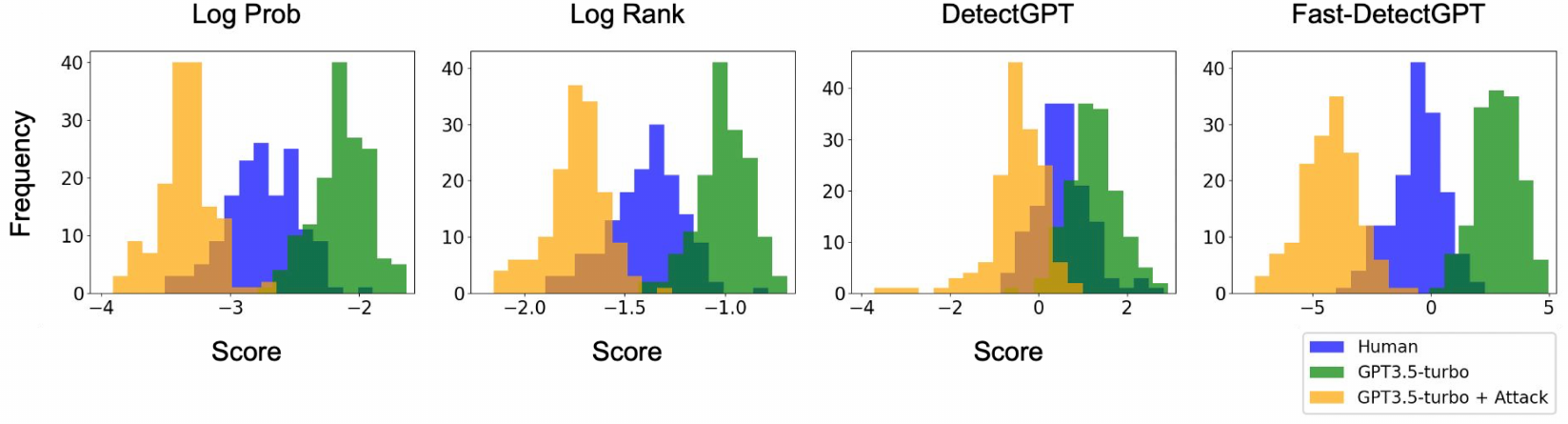}
    \caption{Histograms show the distributions of different detection scores for human-written text, GPT-3.5-Turbo generated text, and \emph{RAFT}-attacked GPT-3.5-Turbo text. The horizontal axis represents the raw output from the detector. The diagrams illustrate that our attack effectively shifts the distribution of generated data towards the negative region, fooling the detectors.}
    \label{fig:hist}
    \vspace{-0.85em}
\end{figure*}
\vspace{-1.5mm}
\section{Related Work}
\textbf{Substitution-based Attacks against NLP targets: } While existing gradient-based adversarial attacks are effective in the vision and speech domains \citep{carlini2017towards}, attacks in the text domain present unique challenges due to its discrete nature. Attacks in the natural language domain are also constrained by language fluency, semantic consistency, and human prediction consistency. \citet{jin2020bert} introduce a black-box word substitution-based attack that fulfills all these criteria by utilizing semantic similarity and POS matching to greedily replace words with synonyms until a successful attack. In this work, we use an LLM instead to generate semantically consistent word replacement candidates and greedily select the POS-consistent word that most effectively attacks the target detector. \\ \\
\textbf{LLM Detector Attack Frameworks: }Existing algorithms for detecting machine-generated text can be categorized into three categories: supervised classifiers \citep{solaiman2019release, hovy2016enemy, zellers2019defending},  watermark detectors \citep{kirchenbauer2023watermark, grinbaum2022ethical, abdelnabi2021adversarial}, and zero-shot statistical-based methods \citep{mitchell2023detectgpt, tian2023gptzero, lavergne2008detecting, solaiman2019release, gehrmann2019gltr, ippolito2019automatic}. As new detection methods continue to be developed, parallel efforts in red-teaming these detectors have also gained momentum. The primary techniques for attacking them include text paraphrasing and word replacement. \citet{krishna2024paraphrasing} present \emph{DIPPER}, a paraphrase generation model that can be conditioned on surrounding context to effectively attack state-of-the-art detectors while controlling output diversity and maintaining semantic consistency. \citet{sadasivan2023can} iterate on this method and present a recursive paraphrasing attack that breaks watermarking and retrieval-based detectors with slight degradation in text quality by using a lightweight T5-based paraphraser model. These attack frameworks are more vulnerable to attack since they are reliant on one large fine-tuned LLM model and utilize smaller paraphrasing models that are relatively weaker than the source LLM. \citet{shi2024red} introduce a word replacement method that utilizes an LLM to randomly generate substitution candidates for multiple words, selecting the optimal replacement candidates using an iterative evolutionary search algorithm to minimize detection score. We improve upon this framework by introducing a replaceable proxy scoring model that uses an auxiliary LLM embedding to rank which words in the machine-generated text should be replaced and greedily select LLM-generated candidates that effectively subvert the target detector.

\section{Method}
\subsection{Preliminaries}
\textbf{Setup: } Given a text passage $\X$ consisting of $N$ words $[x_1,\dots, x_N]$, we consider a black-box detector $D(\X) \in [0,1]$ that predicts whether the input $\X$ is machine-generated or human-written. A higher $D(\X)$ score indicates a greater likelihood that $\X$ is machine-generated. We denote $\tau$ as the detection threshold, such that $\X$ is classified as machine-generated if $D(\X) \geq \tau$. \\ \\ 
\textbf{Adversarial Attack for LLM Detector: } The goal of the attack is to perturb the input passage $\X$ into $\X'$ such that $D(\X')$ incorrectly classifies $\X'$ as human written, while ensuring that $\X'$ remains indistinguishable from human-written text when manually reviewed. To preserve semantic similarity between $\X$ and $\X'$, the number of words to be substituted is constrained to $k\%$ of $\X$. Additionally, to maintain grammatical correctness and fluency, $\x_i$ and $\x'_i$ must have consistent part-of-speech \citep{brill1995transformation}. We formulate the attack on the LLM Detector $\D$ as a constrained minimization problem, with the objective to modify $\X$ such that $D(\X') \leq \tau$:
\begin{equation}
\begin{split}
    \X' = \argmin_{\X'} D(\X') \ \ \  \text{s.t.} \ \ \ pos(\x'_i) = pos(\x_i), \\ \ \ \ \x'_i \in \{\x_i\} \ \cup \ s(\x_i, \X, t),\\ \ \ \ \sum_i^N \mathbf{1}(x'_i \neq x_i) \leq kN  
\end{split}
\end{equation}
where $pos(\x_i)$ returns the part-of-speech label of any word $\x_i$ and $s(\cdot)$ is a word substitution generator that outputs $t$ candidates for $\x_i$ using the surrounding context in $\X$.
\subsection{Our Attack}

\textbf{Finding Important Words for Substitution using a Proxy Task Embedding Objective: } We capitalize on \citet{freestone2024word}'s observations that LLMs share similar latent semantic spaces and perform similarly on semantic tasks. To effectively minimize $D(\X')$, we use a white-box LLM $M$ to perform a word-level task $\F$ that generates a score $\f_i$ for each word that acts as a \textit{proxy }signal for selecting words to replace, where $M$ does not necessarily need to be the same source LLM model used to generate $\X$. We choose LLM embedding tasks correlated with identifying words that would alter the statistical properties of the machine-generated text, such as next-token generation and supervised LLM text detection. From $\F$, we choose 
\begin{equation}
    \X_k = \text{argmax}_{kN}\F(M, \X)
\end{equation}
where $\X_k$ is the subset of $\k\%$ words in $\X$ to perturb from. \\ \\
\textbf{Constraints for Realistic Generation: } To perturb words in $\X_k$ while ensuring that $\X$ remains indistinguishable to a human evaluator as machine-generated, we constrain the replacement words such that they must not induce grammatical errors and are semantically consistent with the original text. We use GPT-3.5-Turbo \cite{openai_gpt35turbo} as our word substitution candidate generator by prompting it with the word to replace and its surrounding context using the following prompt:
\begin{lstlisting}[breakatwhitespace=true, breaklines=true, gobble=0]
Q: Given some input paragraph, we have highlighted a word using brackets. List top {t} alternative words for it that ensure grammar correctness and semantic fluency. Output words only.\n{paragraph}
A: The alternative words are 1. 2. ...
\end{lstlisting}
Using an LLM for word substitution allows us to conveniently obtain context-compatible candidates in one step, instead of needing to compute candidates using word embeddings followed by an additional model to check context compatibility \citep{alzantot2018generating}. After retrieving $t$ replacement candidates from GPT-3.5-Turbo, we filter out words that have inconsistent part-of-speech with the original word by using the NLTK library \citep{nltk} and then select the candidate that minimizes $D$. 
\subsection{Implementation Details}
We set $k$ to 10\% across all experiments to evaluate the effectiveness of our attack with a limited number of changes. We evaluate the effectiveness of using language modeling heads for next-token generation and supervised LLM detection tasks as proxy scoring models to optimally select words to substitute in $\X$. For next-token generation, we use the probability of the next token being $X_{i}$ from the language modeling head as the proxy objective. Intuitively, replacing tokens with the highest likelihood from the LLM allows us to alter the statistical properties of the machine-generated text most effectively. For LLM detection, we iteratively compute the importance of each word based on the decrease in detection score $D(\X)$ by assigning 0 to its corresponding tokens in the detector’s attention mask, and ranking the score changes in descending order, where the word that yields a higher absolute change in detector score is considered to be more important for detection.

\begin{figure*}[h]
    \centering
    \includegraphics[width=\textwidth]{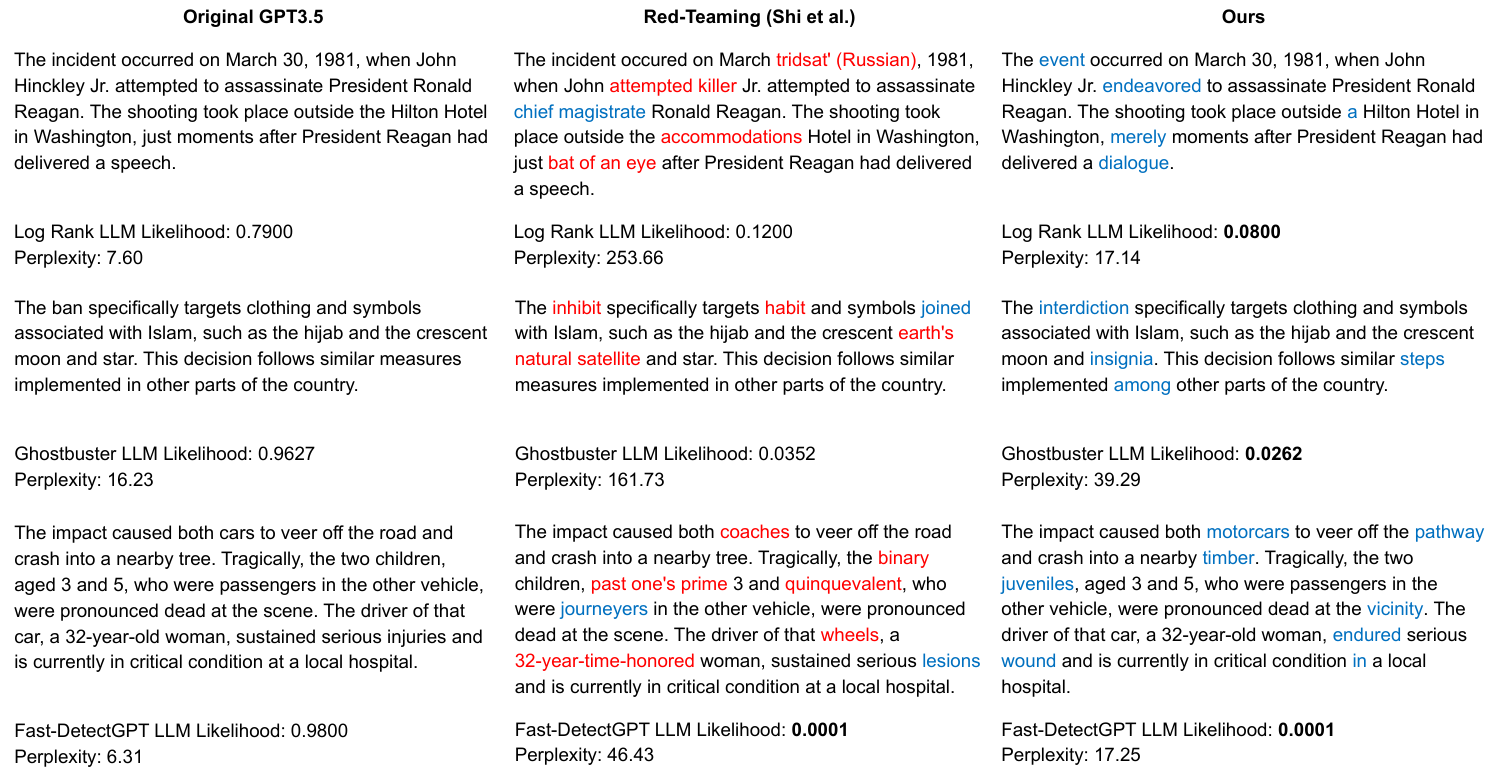}
    \caption{Generated texts from LLMs and their respective attacks using \citet{shi2024red}'s query-based word substitution attack and \emph{RAFT} (ours) using the RoBERTa-large proxy scoring model, evaluated against Log Rank, Ghostbuster, and Fast-DetectGPT detectors. \emph{RAFT} demonstrates the greatest reduction in detection likelihood while maintaining grammatical correctness and semantic consistency with the original text. \textcolor{red}{Red} text represents substituted words with grammatical errors or semantic inconsistencies. \textcolor{blue}{Blue} text represent error-free substitutions.}
    \label{fig:texts}
    \vspace{-1.0em}
\end{figure*}

\section{Experiments}
\subsection{Datasets and Metrics}
\textbf{Datasets}:   
We use three datasets to cover a variety of domains and use cases. We use 200 pairs of human-written and LLM-generated samples from each of the XSum \citep{narayan2018don} and SQuAD \citep{rajpurkar2016squad} datasets generated by \citet{bao2023fast} using GPT-3.5-turbo. Additionally, we use the ArXiV Paper Abstract dataset \citep{mao2024raidar} which contains 350 abstracts generated using GPT-3.5-turbo from ICLR conference papers.
\\ \\
\textbf{Metrics}:   
 We use the Area Under the Receiver Operating Characteristic Curve (AUROC) to summarize detection accuracy for our attack framework under various thresholds. We also measure the True Positive Rate at a 5\% False Positive Rate (TPR at 5\% FPR), as it is imperative in this context for human-written text to not be misclassified as machine-generated text. To measure text quality, we measure the perplexity of the attacked text against GPT-NEO-2.7B \citep{gao2020pile}. 

\subsection{Embeddings for Proxy Scoring Models}
We use the language modeling heads of GPT-2 \citep{radford2019language}, OPT-2.7B \citep{zhang2022opt}, GPT-NEO-2.7B \citep{gao2020pile}, and GPT-J-6B \citep{gpt-j} for next-token generation, and the RoBERTa-base and RoBERTa-large supervised GPT-2 detector models~\citep{solaiman2019release} for LLM detection as proxy tasks to rank which words from the original text to substitute. We present results for OPT-2.7B and RoBERTa-large proxy scoring models in Table~\ref{tab:maintable} and the rest in Table~\ref{tab:my-table}.

\subsection{Detectors}
We evaluate our attack against a variety of target detection methods: \\
\textbf{Log Likelihood}  \citep{gehrmann2019gltr} is a classical threshold-based zero-shot method where passages with higher log probability scores are more likely to have been generated by the target LLM. \\ 
\textbf{Log Rank}  \citep{solaiman2019release} is a classical threshold-based zero-shot method where passages with above average rank are more likely to have been generated by the target LLM. \\ 
\textbf{DetectGPT} \citep{mitchell2023detectgpt}  is a state-of-the-art zero-shot detector that leverages the likelihood of generated texts to perform thresholding for detecting machine-generated text. \\ 
\textbf{Fast-DetectGPT} \citep{bao2023fast} is a state-of-the-art detector that improves upon DetectGPT by introducing conditional probability curvature to underscore discrepancies in word choices between LLMs and humans to improve detection performance and computational speed. \\
\textbf{Ghostbusters} \citep{verma2023ghostbuster} is a state-of-the-art detector that uses probabilistic outputs from LLMs to construct features to train an optimal detection classifier. \\
\textbf{Raidar} \citep{mao2024raidar} is a state-of-the-art detector that uses prompt rewriting and an output's edit distance to gain additional context about the input.

\begin{table*}[h]
    \centering
    \caption{\emph{RAFT} attack results. We evaluate \emph{RAFT} performance against 6 target detectors using GPT-3.5-Turbo generated text from 3 datasets, measuring the detector's performance before and after attack using the AUROC metric. \textbf{Bolded} AUROC results indicate best attack performance. These results show the superiority of our attack compared to benchmarked methods. }
    \begin{adjustbox}{max width=\textwidth}
    \begin{tabular}{l|c|c|c|c|c|c|c}
    \toprule
    Metric & Log Probability & Log Rank &  Ghostbuster & DetectGPT & Fast-DetectGPT & Raidar & Average\\
    \midrule
    XSum / Unattacked & 0.9577 & 0.9584  & 0.6637  & 0.7853 &   0.9903   & 0.7667 &  0.8537  \\
    Dipper & 0.7981 &   0.8080 &  0.7196 &  0.4693 &  0.9610 &  \textbf{0.4667}  & 0.7038   \\
    Query-based Substitution & 0.0481 &  0.0739 &  0.0980 &   \textbf{0.0384} &   0.2308 &   0.6000 &  0.1815 \\
    OPT-2.7B (Ours) & \textbf{0.0035} &  \textbf{0.0069} &  0.0826 &  0.1273 &  \textbf{0.0006} &  0.7000 &  0.1535   \\
    RoBERTa-large (Ours) & 0.0346 &   0.0568  & \textbf{0.0004} &   0.0704 &  0.0371 &   0.6000 &   \textbf{0.1324}  \\
    \addlinespace
    \hline
    \addlinespace
    SQuAD / Unattacked & 0.9027 &  0.9075 &  0.7659  & 0.7916 &  0.9800 &   0.7833 &  0.8552  \\
    Dipper & 0.7929 &  0.8067 &   0.7959 &  0.5916 &  0.9492 &  0.\textbf{5333} &   0.7449  \\
    Query-based Substitution & 0.1542 &   0.1852 &  0.2032 &  0.1408 &  0.3624 &   0.8333 &  0.3132  \\
    OPT-2.7B (Ours) & \textbf{0.0496} &  \textbf{0.0659} &  0.0851 & 0.1539 &  \textbf{0.0131} &  0.8333 &  0.2002  \\
    RoBERTa-large (Ours) & 0.0942 &  0.1199 &  \textbf{0.0166} &  \textbf{0.1262} &   0.1039 &   0.7167 & \textbf{0.1963} \\
    \addlinespace
    \hline
    \addlinespace
    Abstract / Unattacked & 0.6329 &  0.6502 &  0.8455 &  0.1538 &   0.9148 &  0.7667 &   0.6607  \\
    Dipper & 0.5029 &  0.5370 &  0.8826 &  0.1049 & 0.9441 &  0.6833 & 0.6091 \\
    Query-based Substitution & 0.0234 &   0.0364 &  0.3142 &   0.0046 &  0.2976 &  0.7167 &  0.2322  \\
    OPT-2.7B (Ours) & 0.0945 &  0.1249 &   0.0841 &  0.3131 &  \textbf{0.0399} &  0.7667 &   0.2372 \\
    RoBERTa-large (Ours) & \textbf{0.0162} &   \textbf{0.0336} & \textbf{0.0374} &  \textbf{0.0044}  & 0.1481 &   \textbf{0.6500} &  \textbf{0.1666} \\
    \bottomrule 
    \end{tabular}
    \end{adjustbox}
    \label{tab:maintable}
\end{table*}

\begin{table*}[h]
    \centering
    \small
    \caption{Perplexity of text after different attacks measured by GPT-NEO-2.7B. \emph{RAFT} attacked texts were optimized against Fast-DetectGPT detector. Lower perplexity indicates better text quality. The results show that our attack is able to maintain text quality while subverting detection.}
    \begin{tabular}{l|ccccc}
        \toprule
        Dataset & Unattacked & Dipper & Query-based Substitution & OPT-2.7B (Ours) & RoBERTa-large (Ours) \\
        \midrule
        XSum & 8.4804 & 11.3649 & 28.0979 & 17.6181 & 22.4542 \\
        SQUAD & 9.7947 & 11.9064 & 30.0879 & 19.6190 & 25.1480 \\
        Abstract & 12.9136 & 15.2685 & 36.6523 & 26.8810 & 31.6123 \\
        \bottomrule
    \end{tabular}
    \label{tab:models}
\end{table*}
\subsection{Baselines}
We compare our attack method with \emph{DIPPER} \citep{krishna2024paraphrasing}, a paraphrase generation model using settings of 20 lexical diversity and 60 order diversity. This corresponds to about 20\% lexical modification – the minimum modification we can set on this method. We also compare with \citet{shi2024red}'s query-based word substitution attack and limit the number of substituted words to be at most 10\% to match our substitution frequency.
\subsection{Results}
\begin{figure*}[h]
    \centering
    \includegraphics[width=\textwidth]{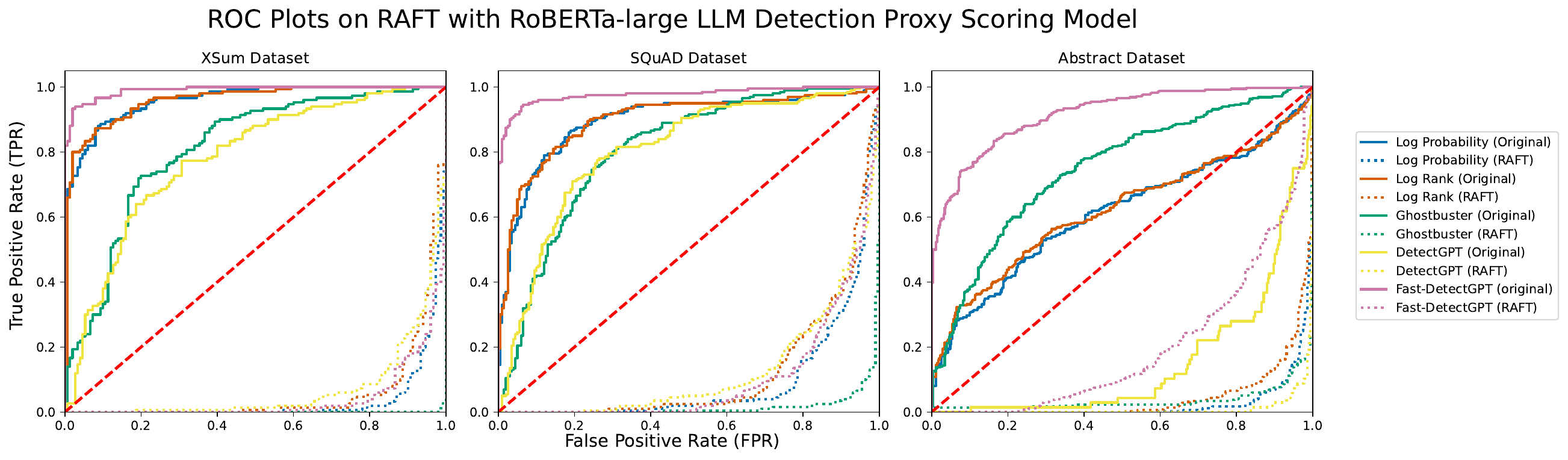}
    \caption{ROC curves for machine-generated XSum dataset under \emph{RAFT} attack, using RoBERTa-large as the proxy scoring model, compared across multiple detectors. The ROC curves for both attacked and unattacked text provide a more comprehensive evaluation of \emph{RAFT}'s robustness in subverting text detectors than single metrics.}
    \label{fig:roc-roberta-large}
\end{figure*}

Tables~\ref{tab:maintable} and~\ref{tab:models} demonstrate that our attack effectively compromises all tested detectors while causing only a modest change in perplexity from the original machine-generated text. Using next-token generation with OPT-2.7B and LLM detection with RoBERTa-large as proxy scoring models for \emph{RAFT} achieved lower AUROC across all datasets and target detectors when compared to the original text, and in most cases, lower than both \emph{DIPPER} and \citet{shi2024red}'s query-based word substitution. Although \emph{DIPPER} preserves the text quality better in terms of perplexity, its AUROC is significantly higher than \emph{RAFT} and \citet{shi2024red}'s attack. The TPR at 5\% FPR was 0 for almost all \emph{RAFT} attacked text, which we present in Table~\ref{tab:tpr_fpr}. For more insight, we present the ROC curve for our experiments in Figure \ref{fig:roc-roberta-large}. Between \citet{shi2024red}'s query-based attack and our method, \emph{RAFT} consistently yields lower perplexity scores across all scenarios. Raidar stands out as the most robust detector against attacks, likely due to the unique edit distance of rewriting used in the approach. Qualitative results shown in Figures~\ref{fig:teaser} and \ref{fig:texts} highlight our method's semantic consistency and language fluency. Additionally, cosine similarity calculations between the original and perturbed texts shown in Table \ref{tab:Cosine Similarity} using state-of-the-art LLMs Mistral-7B-v0.3~\cite{jiang2023mistral} and Llama-3-8B \cite{dubey2024llama} highlight their strong semantic similarity. We also show in Figure~\ref{fig:hist} that our attack effectively alters the distribution of detection likelihood scores, diverging from the distribution associated with the machine-generated text, thereby subverting detection. 
\begin{table}
    \centering
    \small
    \caption{Cosine similarity, evaluated across multiple LLM embeddings between the original texts and those perturbed by \emph{RAFT} using RoBERTa-base as the proxy scoring model and Fast-DetectGPT as the target detector, indicates that the texts maintain semantic similarity.}
    \begin{tabular}{l|ccc}
        \toprule
        Embedding Model & XSum & SQuAD & Abstract \\
        \midrule
        RoBERTa-large & 0.9999 & 0.9999 & 0.9999 \\
        Llama-3-8B & 0.9747 & 0.9759 & 0.9841 \\ 
        Mistral-7B-v0.3 & 0.9761 & 0.9735 & 0.9847 \\ 
        \bottomrule
    \end{tabular}
    \label{tab:Cosine Similarity}
    \vspace{-0.9em}
\end{table}

\begin{table*}[h]
\centering
\small
\vspace{-1.0em}
\caption{AUROC of \emph{RAFT}-attacked text, using Word2Vec embedding model trained on the Google News corpus for word replacement candidate generation instead of GPT-3.5-turbo on the XSum and Abstract datasets, suggests that using a classic word embedding in place of an LLM also yields effective results.}
\begin{tabular}{l|cccc}
\toprule
        Proxy Model/Detector & Log Probability & Log Rank & Ghostbuster & Fast-DetectGPT \\
        \midrule
        XSum/Unattacked & 0.9577 & 0.9584 & 0.6637 & 0.9903\\
        OPT-2.7B (Ours) & 0.0052 & 0.0144 & 0.0408 & 0.0034\\ 
        RoBERTa-large (Ours) & 0.0016 & 0.0064 & 0.0000 & 0.0698 \\
        \addlinespace
        \hline
        \addlinespace
        Abstract/Unattacked & 0.6329 & 0.6502 & 0.8455 & 0.9148\\
        OPT-2.7B (Ours) & 0.1041 & 0.1577 & 0.0873 & 0.0711 \\
        RoBERTa-large (Ours) & 0.0021 & 0.0040 & 0.0075 & 0.0346\\ 
        \bottomrule
\end{tabular}
\label{tab:vector-embedding}
\end{table*}

\subsection{Human Evaluation}
To validate that \emph{RAFT} preserves text quality, we conducted a crowd-sourced human evaluation using Amazon Mechanical Turk (MTurk). We selected the first 100 pairs of human-written and GPT-3.5-Turbo-generated texts from the XSum, SQuAD, and Abstract datasets. After applying \emph{RAFT} to the LLM-generated text, three MTurk workers evaluated each pair of original human-written and \emph{RAFT}-modified texts, indicating their preference for one of them or expressing no preference. \emph{RAFT}'s perturbations were deemed indistinguishable from human-written text if two or more annotators either preferred the perturbed text or were indifferent. To ensure English proficiency, we included a screening question using a text comparison task sourced from the ETS TOEFL website. Out of valid 396 responses, 185 preferred the human-written texts, 182 were indifferent, and 29 responses were excluded for rating both texts as low quality. A two-tailed binomial test yielded a p-value of 0.917 at $\alpha < 0.05$, supporting the null hypothesis that the two texts are indistinguishable. The Fleiss' kappa was 0.774, indicating strong agreement among annotators.

\section{Discussion}
\subsection{\textbf{Effect of Scoring Model}}
We perform ablation studies to evaluate the isolated effectiveness of the proxy scoring model (ranking) and the greedy selection of generated POS-consistent replacement words aimed at subverting detection (optimization). For brevity, we refer to these two methods as "ranking" and "optimization", respectively. As shown in Table~\ref{tab:scoring model ablation}, the study is conducted under four settings: neither ranking nor optimization, ranking only, optimization only, and both ranking and optimization. The results indicate that ranking is about as effective as optimization, significantly reducing AUROC  when applied, supporting the idea that LLM embeddings are transferable. However, the effects of ranking and optimization are not necessarily additive. 
\begin{table*}[h]
    \centering
    \small
     \caption{Effect of scoring models. To show the isolated effectiveness of the proxy scoring model, we attack the Ghostbuster and Fast-DetectGPT detectors using OPT-2.7B next-token generation on the XSum dataset across four different configurations. Here, "ranking" refers to the proxy scoring model, and "optimization" refers to the greedy selection of generated POS-consistent words against the target detector. The results indicate both techniques are effective, but their combined effect is not necessarily additive. \textbf{Bolded} AUROC results denote best attack performance.}
    \begin{tabular}{l|c|c|c|c}
        \toprule
        Setting & \multicolumn{1}{c|}{Neither} & \multicolumn{1}{c}{Optimization Only} & \multicolumn{1}{|c}{Ranking Only} & \multicolumn{1}{|c}{Ranking + Optimization}\\
        \midrule
        Ghostbuster & 0.3341 & 0.1001 & \textbf{0.0981} & 0.1000 \\
        Fast-DetectGPT & 0.7510 & 0.0026 & 0.0030 & \textbf{0.0006} \\
        Raidar & 0.7667 & 0.6333 & 0.6667 & \textbf{0.6000} \\
        \bottomrule
    \end{tabular}
    \label{tab:scoring model ablation}
\end{table*}

\begin{table*}[h]
    \centering
    \small
    \caption{Effect of applying our grammar constraint. We use OPT-2.7B and RoBERTa-large as scoring models to attack the Fast-DetecGPT detector on the XSum dataset, comparing performance with and without the part-of-speech constraint on the generated replacement words. The results show a marginal decrease in attack performance but a significant improvement in perplexity when POS constraints are enforced.}
    \begin{tabular}{l|cc|cc}
        \toprule
        & \multicolumn{2}{c|}{No POS Correction} & \multicolumn{2}{c}{POS Correction} \\
        & AUROC & Perplexity & AUROC & Perplexity \\
        \midrule
        OPT-2.7B & 0.0000 & 28.32 & 0.0006 & 17.53 \\
        RoBERTa-large & 0.0062 & 31.36 & 0.0471 & 25.08 \\
        \bottomrule
    \end{tabular}
    \label{tab:pos}
\end{table*}

\begin{figure*}[ht!]
    \centering
    \begin{subfigure}[b]{0.49\textwidth}
        \centering
        \includegraphics[width=\textwidth]{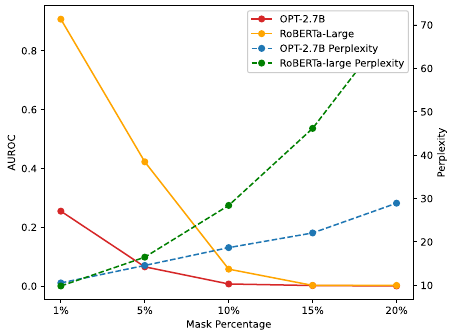}
        \caption{Log Rank}
        \label{fig:img1}
    \end{subfigure}
    \begin{subfigure}[b]{0.49\textwidth}
        \centering
        \includegraphics[width=\textwidth]{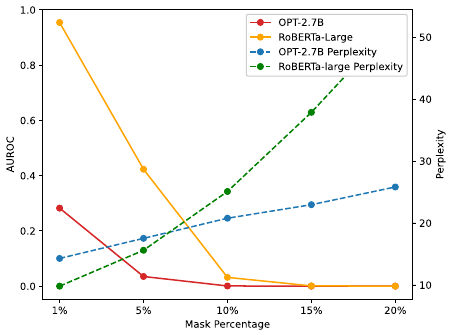}
        \caption{Fast-DetectGPT }
        \label{fig:img2}
    \end{subfigure}
    \caption{Study on the impact of different mask percentages. We use OPT-2.7B and RoBERTa-large as proxy scoring models to attack Log Rank and Fast-DetectGPT detectors on the XSum dataset at 1\%, 5\%, 10\%, 15\%, and 20\% masking rates while measuring detection performance and text quality in terms of AUROC and perplexity. The AUROC approaches 0 at around 10\% with a moderate increase in perplexity. Masking percentages beyond 15\% degrade text quality across both detectors.}
    \vspace{-1.0em}
    \label{fig:masking percentage}
\end{figure*}
\begin{figure*}[h]
    \centering
    \includegraphics[width=0.95\textwidth]{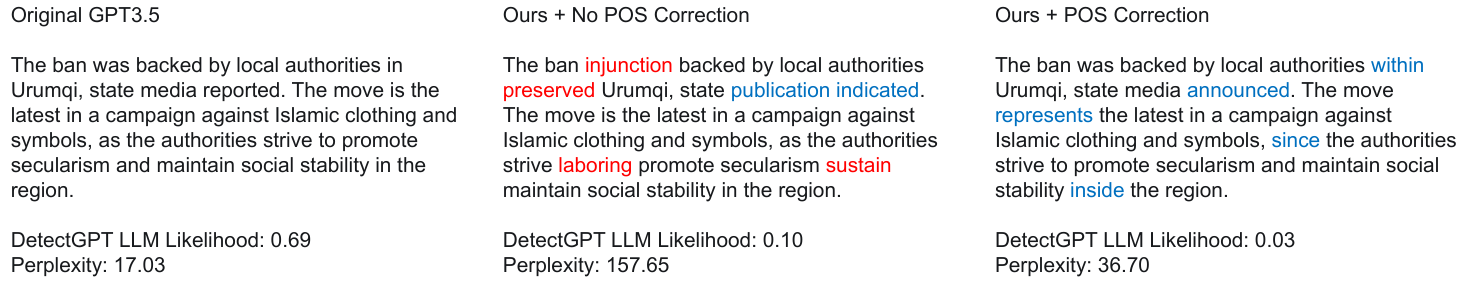}
    \vspace{-0.5em}
    \caption{Comparison on the effects of POS tagging. On the left is unmodified text generated by GPT3.5-turbo; in the middle is \emph{RAFT} attacked text but without POS consistency constraints; and on the right is \emph{RAFT}  attacked text with POS consistency. This example illustrates that POS tagging significantly enhances text quality both qualitatively and quantitatively as measured by perplexity, without compromising detection performance.}
    \vspace{-1.0em}
    \label{fig:pos}
\end{figure*}

\subsection{\textbf{Impact of Word Replacement Generation Method}}
We evaluate the effectiveness of replacing GPT-3.5-Turbo with a traditional Word2Vec embedding model \cite{mikolov2013efficient}. Specifically, we use the Word2Vec model trained on Google News corpus, which contains 1 billion words \cite{mikolov2013distributed}, to locally retrieve $t=10$ POS-consistent synonyms as word replacement candidates. We show in Table \ref{tab:vector-embedding} that using a word embedding model instead of an LLM also produces effective results. 

\subsection{\textbf{Impact of Masking Percentage}}
We evaluate the performance and text quality of \emph{RAFT} across various masking percentages. Figure \ref{fig:masking percentage} shows that the AUROC stabilizes around 0 when the masking percentage reaches 10\%, accompanied by a moderate increase in perplexity. Masking percentages exceeding 15\% are unnecessary and lead to a significant degradation in text quality.

\subsection{\textbf{Impact of the Source Generation Model}}
\begin{figure*}[h!]
    \centering
    \includegraphics[width=0.95\textwidth]{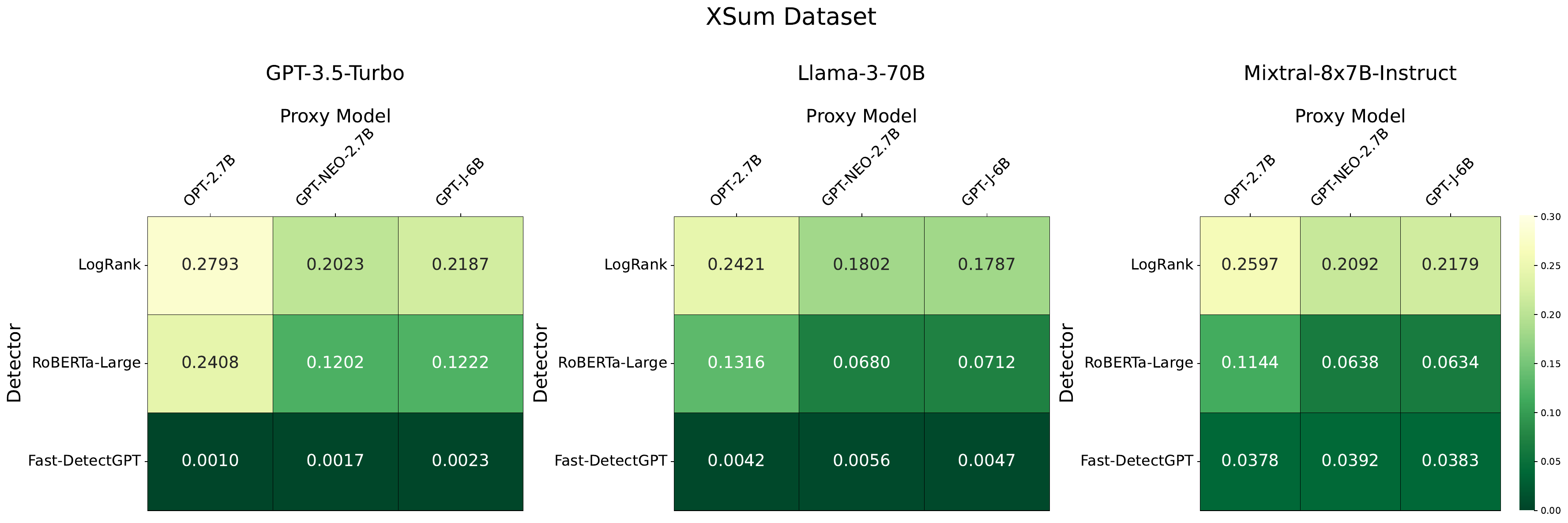}
    \newline
    \includegraphics[width=0.95\textwidth]{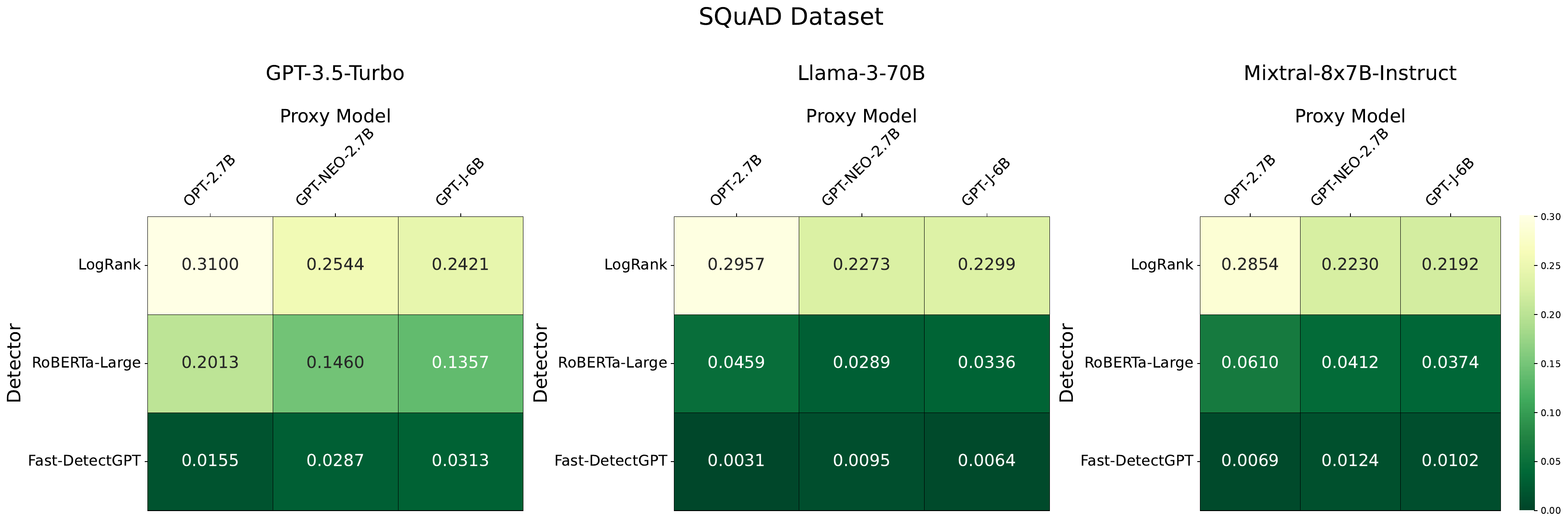}
    \caption{AUROC of using \emph{RAFT} under various source generation models and proxy scoring models against Log Rank, RoBERTa-large, and Fast-DetectGPT detectors on the XSum and SQuAD dataset. The results demonstrate \emph{RAFT} remains robust under various source generation models and proxy scoring model pairs.}
    \label{tab:new_XSum}
\end{figure*}

We study the effectiveness of \emph{RAFT} under different source generation models. We evaluate its effectiveness on text generated using GPT-3.5-Turbo, Llama-3-70B \citep{touvron2023llama}, and Mixtral-8x7B-Instruct \citep{jiang2024mixtral}, which represent a set of LLMs of varying size, architecture, and trained corpora. We utilize the same generation parameters as those employed by \cite{bao2023fast} for producing the GPT-3.5-turbo generated XSum and SQuAD datasets for Llama-3-70B and Mixtral-8x7B-Instruct. We show in Figure~\ref{tab:new_XSum} that \emph{RAFT} remains highly effective when next-token generation is used as a proxy task with OPT-2.7B, GPT-NEO-2.7B, GPT-J-6B embeddings model for subverting detection against Log Rank, RoBERTa-large, and Fast-DetectGPT detectors.

\begin{table*}[t!]
    \centering
    \small
    \caption{Adversarial training results. We train the Raidar detector on texts with and without word swapping, denoted as Training Method, and evaluated its performance on samples with (Attack) and without (Clean) word swapping. The result shows the detector becomes more robust under adversarial training. \textbf{Bolded} AUROC results denote highest-performing detector.}
    \begin{adjustbox}{max width=0.95\textwidth}
    \begin{tabular}{l|cc|cc|cc}
    \toprule
    Dataset & \multicolumn{2}{|c|}{XSum} & \multicolumn{2}{|c}{SQuAD} & \multicolumn{2}{|c}{Abstract} \\
    Training Method & Normal & Adversarial & Normal & Adversarial & Normal & Adversarial \\
    \midrule
         Clean AUROC & \textbf{0.8000} & 0.7500 & \textbf{0.6833} & 0.6667 & 0.6500 & \textbf{0.6833} \\
         Attack AUROC & 0.6000 & \textbf{0.7333} & 0.7167 & \textbf{0.8000} & 0.6500 & \textbf{0.7167} \\
    \bottomrule
    \end{tabular}
    \end{adjustbox}
    \label{tab:adversarial training}
    \vspace{3em}
\end{table*}

\subsection{\textbf{Transferrability of \emph{RAFT} on other Detectors}}
We study the transferability of \emph{RAFT}-attacked text across various detectors. We evaluate the attacked text generated by using OPT-2.7B next-token generation and RoBERTa-large LLM detection proxy scoring tasks optimized against LogRank and DetectGPT detectors on GhostBuster and Fast-DetectGPT. The results, presented in Table~\ref{tab:transferability}, show that the AUROC only decreases slightly, suggesting that our attack is highly transferable. 

\subsection{\textbf{\emph{RAFT} for Adversarial Training}}
We present evidence that our attack not only effectively subverts detectors but can also enhance their robustness through adversarial training. As shown in Table~\ref{tab:adversarial training}, after the Raidar detector undergoes adversarial training on \emph{RAFT}-attacked text, it consistently demonstrates a significant increase in detection performance under attack compared to the performance decrease observed before retraining. For the Abstract dataset, the AUROC for both attacked and non-attacked text samples increases, indicating that \emph{RAFT} can enhance the robustness of existing detectors through adversarial training. We present this as an important direction for future research.

\section{Conclusion}
\vspace{-2mm}
We introduce \emph{RAFT}, an adversarial attack framework for subverting machine-generated text detectors by leveraging auxiliary LLM embeddings. Our method effectively identifies optimal words to perturb using a proxy LLM embedding and perturbs them such that the original text remains semantically consistent, grammar error-free, and reads fluently. Experimental results and manual annotation exercises show that our method successfully compromises various LLM detection methods while maintaining text quality and semantic consistency, highlighting the need for robust LLM content detectors. We also demonstrate that the outputs from \emph{RAFT} can be used to enhance the resilience of existing detectors through adversarial training.

\section{Limitations}
While we demonstrate \emph{RAFT}'s effectiveness in compromising various LLM detectors, there are several limitations to note: \\ \\
\textbf{Scalability of Human Evaluations}: While our manual human evaluation study demonstrated that \emph{RAFT}'s perturbations are realistic and are not necessarily less preferred from the original text, larger-scale human evaluations are necessary to validate the quality and realism of the perturbed texts robustly. Furthermore, we did not extensively explore the demographic and linguistic backgrounds of the human evaluators, which may induce bias in our study.  \\ \\
\textbf{Computational \& Cost Overhead}: The runtime performance of \emph{RAFT} is shown in Table \ref{tab:runtime}. Generating substitution candidates using GPT-3.5-Turbo or using a word embedding for each selected candidate replacement word introduces significant computation and cost overhead. This may limit the practicality of this attack in real-time or in budget-constrained environments. Developing more efficient prompting strategies for effective word-level substitutions would be essential for practical use. \\ \\
\textbf{Fixed Perturbation Rate}: We fixed the perturbation rate at 10\% across all experiments, which is less than the rate set in \citet{shi2024red} and \citet{krishna2024paraphrasing}. While this provides a consistent and strong benchmark, it does not account for scenarios where a smaller perturbation rate may be more effective. Exploring adaptive perturbation strategies based on text complexity and detection sensitivity may yield a more efficient and effective attack. \\ \\
\textbf{Limited Detector Evaluation}: \emph{RAFT} was tested against various types of LLM detectors. However, as new detection methods emerge, we must continuously evaluate our attack's robustness on novel approaches.

\section{Ethics Statement}
\vspace{-2mm}
While our paper presents a method to subvert detection of machine-generated text by LLM detectors, it is imperative to acknowledge that LLMs are predominantly utilized in good faith and have a wide variety of benefits to society, such as improving one's work and efficiency. By scrutinizing LLM detectors through red-teaming, we highlight current vulnerabilities in these systems and urgently advocate for the development of more resilient mechanisms. While we introduce how examples generated by \emph{RAFT} can be utilized for adversarial training, future work should emphasize the development of robust defense mechanisms.

\section{Acknowledgements}
This work was supported in part by multiple Google Cyber NYC awards, Columbia SEAS/EVPR Stimulus award, and Columbia SEAS-KFAI Generative AI and Public Discourse Research award.
\bibliography{custom}
\onecolumn
\appendix
\counterwithin{table}{section}

\section{Appendix}

\subsection{Tables}

\label{sec:appendix}

\begin{table*}[h]
    \centering
    \caption{Our attack results using additional proxy score models demonstrate \emph{RAFT} is effective against various target detectors, scoring similarly to results shown in Table~\ref{tab:maintable}. GPT-2, GPT-NEO-2.7B, and GPT-6B use next token generation and RoBERTa-base uses LLM detection as proxy scoring model tasks. Metric reported is AUROC.}
    
    \label{tab:my-table}
    \begin{adjustbox}{max width=\textwidth}
    \begin{tabular}{l|c|c|c|c|c|c}
    \toprule
    Dataset / Method & Log Probability & Log Rank & Ghostbuster  & DetectGPT & Fast-DetectGPT & Average \\
    \midrule
    XSum / Unattacked & 0.9577 & 0.9584 & 0.6637 &  0.9903 & 0.8925 & 0.8925 \\
    GPT-2 & 0.0046 & 0.0196 & 0.0453 &  0.0211 & 0.0182 & 0.0217 \\
    GPT-NEO-2.7B & 0.0052 & 0.0144 & 0.0408 &  0.0120 & 0.0160 & 0.0177 \\
    GPT-6B & 0.0034 & 0.0156 & 0.0426 &  0.0202 & 0.0160 & 0.0196 \\
    RoBERTa-base & 0.0372 & 0.0584 & 0.0003 &  0.0621 & 0.0390 & 0.0394 \\
    \addlinespace
    \hline
    \addlinespace
    SQuAD / Unattacked & 0.9027 & 0.9075 & 0.7659 & 0.9800 & 0.8890 & 0.8890 \\
    GPT-2 & 0.0595 & 0.0959 & 0.0862 &  0.0574 & 0.0695 & 0.0737 \\
    GPT-NEO-2.7B & 0.0532 & 0.0839 & 0.0831 &  0.0518 & 0.0617 & 0.0667 \\
    GPT-J-6B & 0.0524 & 0.0883 & 0.0667 & 0.0508 & 0.0600 & 0.0636 \\
    RoBERTa-base & 0.0999 & 0.1262 & 0.0175 & 0.1433 & 0.1068 & 0.0988 \\
    \addlinespace
    \hline
    \addlinespace
    Abstract / Unattacked & 0.6329 & 0.6502 & 0.8455 & 0.9148 & 0.7609 & 0.7609 \\
    GPT-2 & 0.1466 & 0.1960 & 0.0912 &  0.1885 & 0.1353 & 0.1515 \\
    GPT-NEO-2.7B & 0.1041 & 0.1577 & 0.0873 &  0.1491 & 0.1050 & 0.1206 \\
    GPT-J-6B & 0.1066 & 0.1624 & 0.0794 & 0.1515 & 0.1075 & 0.1215 \\
    RoBERTa-base & 0.0296 & 0.0426 & 0.0388 &  0.0079 & 0.1994 & 0.0637 \\
    \addlinespace
    \bottomrule
    \end{tabular}
    \end{adjustbox}
    \label{tab:appendix}
\end{table*}
\begin{table*}[h]
    \centering
     \caption{Performance of \emph{RAFT} attack measured by TPR at 5\% FPR. Our results show that \emph{RAFT} significantly lowers the TPR at 5\% FPR to nearly 0 across all detectors and datasets, highlighting the robustness of our approach.}
\begin{adjustbox}{max width=\textwidth}
    \begin{tabular}{l|c|c|c|c|c}
    \toprule
    Metric & Log Probability & Log Rank &  Ghostbuster & DetectGPT & Fast-DetectGPT \\
    \midrule
    XSum / Unattacked & 0.7800 & 0.8067  & 0.2200  & 0.1667 &   0.9400  \\
    OPT-2.7B (Ours) & 0.0000 & 0.0000  & 0.0000  & 0.0000 &   0.0000  \\
    RoBERTa-base (Ours)  & 0.0000 & 0.0000  & 0.0000  & 0.0000 &   0.0000  \\
    RoBERTa-large (Ours)  & 0.0000 & 0.0000  & 0.0000  & 0.0000 &   0.0000  \\
    \addlinespace
    \hline
    \addlinespace
    SQuAD / Unattacked & 0.5750 & 0.6050  & 0.1650  & 0.1533 &   0.9150  \\
    OPT-2.7B (Ours) & 0.0000 & 0.0000  & 0.0000  & 0.0000 &   0.0150  \\
    RoBERTa-base (Ours)  & 0.0000 & 0.0000  & 0.0000  & 0.0000 &   0.0000  \\
    RoBERTa-large (Ours)  & 0.0000 & 0.0000  & 0.0000  & 0.0000 &   0.0000  \\
    \addlinespace
    \hline
    \addlinespace
    Abstract / Unattacked & 0.2086 & 0.2257  & 0.2314  & 0.0000 &   0.6600  \\
    OPT-2.7B (Ours) &  0.0000 & 0.0000  & 0.0200  & 0.0000 &   0.0229  \\
    RoBERTa-base (Ours)  & 0.0000 & 0.0000  & 0.0171  & 0.0000 &   0.0000  \\
    RoBERTa-large (Ours)  & 0.0000 & 0.0000  & 0.0143  & 0.0000 &   0.0000  \\
    \bottomrule 
    \end{tabular}
    \end{adjustbox}
    \label{tab:tpr_fpr}
\end{table*}
\clearpage
\begin{table*}[h]
    \centering
     \caption{Transferability of \emph{RAFT} attacked text. We evaluate \emph{RAFT} perturbed text, using OPT-2.7B and RoBERTa-large proxy scoring models against LogRank and DetectGPT detectors, on LogRank, GhostBuster, DetectGPT, and Fast-DetectGPT detectors. AUROC metrics show only a slight decrease, suggesting our attack is highly transferable. }
    \begin{adjustbox}{max width=\textwidth}
    \begin{tabular}{l|ccc|ccc}
    \toprule
    \emph{RAFT}-optimized Detector & \multicolumn{3}{|c|}{Log Rank} & \multicolumn{3}{|c}{DetectGPT} \\
    \emph{RAFT} Proxy Score Model / Transfer Detector & GhostBuster & DetectGPT & Fast-DetectGPT & Log Rank & GhostBuster & Fast-DetectGPT \\
    \midrule
         OPT-2.7B & 0.1082 & 0.1411 & 0.0022 & 0.0235 & 0.1264 & 0.0059  \\
         RoBERTa-large & 0.0578 & 0.0498 & 0.1541 & 0.2247 & 0.1116 & 0.2927 \\
    \bottomrule
    \end{tabular}
    \end{adjustbox}
    \label{tab:transferability}
\end{table*}
\begin{table*}[h]
\small
    \centering
     \caption{Evaluation of \emph{RAFT} by using higher-performing LLMs, based on MMLU benchmark score \cite{MMLU}, for next-token generation as proxy scoring model on the XSum dataset.  GPT-4o \cite{openai_hello_gpt4o} is used for word replacement candidate generation instead of GPT-3.5-Turbo. The results illustrate that \emph{RAFT} is highly effective on more recent models.}
    \begin{adjustbox}{max width=\textwidth}
    \begin{tabular}{l|cc}
    \toprule
    Proxy Scoring Model & AUROC & TPR at 5\% FPR \\
    \midrule
    XSum / Unattacked & 0.9903 & 0.9400 \\
    Llama-3-8B & 0.0485 & 0.0000 \\
    Mistral-7B-v0.3 & 0.2071 & 0.0000 \\
    Phi-2-2.7B \cite{bubeck2023phi} & 0.1873 & 0.0000 \\
    \bottomrule
    \end{tabular}
    \end{adjustbox}
    \label{tab:new_models}
\end{table*}

\subsection{Human Evaluation Task Details}
The workers were paid \$0.05 USD for each
example. The annotation time for each example
varies, but the estimated wage rate is \$9/hour,
which is higher than the US minimum wage (\$7.25/hour). 
\\ \\ MTurk Task Prompt:

\lstset{escapeinside={(*@}{@*)}}
\begin{lstlisting}%[breakatwhitespace=true, breaklines=true, gobble=0]
Prompt: Please compare the 2 texts below. Which one is more fluent?
Text 1: ${Text 1}
Text 2: ${Text 2}

Options:
(*@\textbullet@*)  Text 1 is better
(*@\textbullet@*)  Text 2 is better
(*@\textbullet@*)  No Preference
(*@\textbullet@*)  Both texts are equally bad
\end{lstlisting}
Note that \$\{Text 1\} and \$\{Text 2\} are shuffled between the original human-written text and \emph{RAFT} perturbed text to avoid selection bias.  

\subsection{\emph{RAFT} Runtime Performance}

\begin{table*}[ht]
    \caption{We execute \emph{RAFT} on a Linux compute cluster equipped with 188 GB of RAM and an NVIDIA A100 GPU with 40 GB of memory. Using RoBERTa-base as the proxy scoring model and Fast-DetectGPT as the target detector, both loaded on the GPU, we run \emph{RAFT} on the XSum dataset. Word replacement candidates are generated using GPT-3.5-Turbo.}
    \centering
    \small
    \begin{tabular}{p{1.2cm} p{2.3cm} p{2.5cm} p{2.7cm} p{2.4cm} p{2.5cm}}
    \toprule
    \textbf{No. Samples} & \textbf{Masking Rate (k\%)} & \textbf{Avg. No. Words / Sample} & \textbf{Avg. No. Words Replaced / Sample}  & \textbf{Avg. Runtime (s) / Sample} & \textbf{Avg. Runtime (s) / Word Replaced} \\
    \midrule
    150 & 10\% & 181 & 18 & 21.64 & 1.20 \\
    \bottomrule
    \end{tabular}
    \label{tab:runtime}
\end{table*}

\end{document}